\documentclass[letterpaper, 10 pt, conference]{ieeeconf}  

\IEEEoverridecommandlockouts                              

\usepackage{graphicx}
\usepackage{subcaption}
\usepackage{amsmath} 
\usepackage{amssymb}  
\usepackage{cite}

\title{\LARGE \bf
Collective Decision-Making on Task Allocation Feasibility
}

\author{Samratul Fuady, Danesh Tarapore, Shoaib Ehsan, and Mohammad D. Soorati 
\thanks{School of Electronics and Computer Science, University of Southampton, Southampton, SO17 1BJ, UK
        {\tt\small \{s.fuady, d.s.tarapore, s.ehsan, m.soorati\}@soton.ac.uk}}%
}

\begin{document}

\maketitle
\thispagestyle{empty}
\pagestyle{empty}

\begin{abstract}

Robot swarms offer the potential to bring several advantages to the real-world applications but deploying them presents challenges in ensuring feasibility across diverse environments. Assessing the feasibility of new tasks for swarms is crucial to ensure the effective utilisation of resources, as well as to provide awareness of the suitability of a swarm solution for a particular task. In this paper, we introduce the concept of distributed feasibility, where the swarm collectively assesses the feasibility of task allocation based on local observations and interactions. We apply Direct Modulation of Majority-based Decisions as our collective decision-making strategy and show that, in a homogeneous setting, the swarm is able to collectively decide whether a given setup has a high or low feasibility as long as the robot-to-task ratio is not near one.

\end{abstract}

\section{INTRODUCTION}

Evaluating the feasibility of new tasks for robot swarms, i.e., whether a group of robots can complete a set of assigned tasks, is essential to ensure that they can effectively and safely execute the assigned tasks in real-world scenarios. When deploying a swarm to cover dangerous substances in an area before human entry, it is crucial to first verify the swarm's capability to accomplish the task. Skipping this verification can lead to situations where the swarm is unable to effectively execute the assigned tasks, potentially causing harm to the robots, the surrounding environment, and humans. While the feasibility evaluation of tasks has been discussed in single robot systems, e.g., manipulator robots \cite{Bouhsain2023} and humanoid robots \cite{Han2020}, the discussion in multi-robot systems or swarms remains very limited. 
In many cases, tasks are made feasible for the swarm by relaxing requirements \cite{Mayya2021}, prioritizing tasks \cite{Notomista2022}, over-provisioning robots \cite{Valentini2022, Lindsay2021}, or enabling sequential task execution \cite{Khaluf2020}. In centralized multi-robot task allocation, feasibility is often integrated into the optimization problem \cite{Choudhury2022, Fu2023, Gosrich2023, Martin2023}. As swarms inherently operate through local interaction \cite{Dorigo2021}\cite{hamann2018swarm}, the evaluation of feasibility needs to be conducted in a decentralized manner. Therefore, we introduce the concept of distributed feasibility, where the swarm collectively decides whether a set of tasks is feasible based on robots' local observations and interactions. This differs from other research on evaluating feasibility in swarm \cite{Gu2023}, where feasibility is centrally calculated using predictive formal modeling, resulting in heavy resource usage that is not scalable.

In this paper, we consider the problem of determining the feasibility of a set of tasks by a swarm of robots. We assume that the robots can sample an area of the task and reach a consensus on whether they should attempt to perform the task if the task is feasible with the given resource. A set of tasks is called feasible if all tasks can be completed by the robots in the swarm (no time limit). We adopt the concept of collective decision-making between two features, namely Direct Modulation of Majority-based Decisions (DMMD) as presented in \cite{Valentini2016}\cite{valentini2016b}, with robots and tasks being our features. We investigate the performance of collective decision-making for feasibility through simulation.

\section{COLLECTIVE DECISION-MAKING ON FEASIBILITY}

We consider a homogeneous setting where all robots possess identical capabilities, and tasks have uniform requirements, specifically necessitating one robot for each task. The feasibility requirement translates to having the number of robots $n$ greater than or equal to the number of tasks $m$ ($n \geq m$). As each robot only has access to local information and observations, they will need to collectively decide whether the tasks are feasible for them or not. The problem is similar to the collective perception problem in \cite{Valentini2016}, but in our setting, one of the features, i.e., the robots, is continuously moving at each time step. We modified the opinion quality calculation in DMMD to facilitate collective decision-making on feasibility.

The robots cycle through two distinct behavioral states as part of their decision-making process: exploration state and dissemination state. At the beginning, each robot randomly selects an opinion on whether the task is feasible or not, assigning 0 to \textit{infeasible} and 1 to \textit{feasible} tasks. During the exploration state, the robot counting the tasks $m_{obs}$ and other robots $n_{obs}$ in its vicinity. Based on this data, the robot calculates the quality of its opinion, denoted as $\rho_i$. The quality estimates of opinion 1 (feasible) is given by:
\begin{equation}
    \rho_1(t) = \frac{n_{\text{obs}}(t)}{n_{\text{obs}}(t) + m_{\text{obs}}(t)},
\end{equation}
where $n_{\text{obs}}(t)$ and $m_{\text{obs}}(t)$ represent the number of robots and tasks observed by the robot during the exploration phase. These values are set to 0 at the beginning of the exploration state. The quality of opinion 0 (infeasible) is calculated in similar way.  After the exploration period ends, the robot moves to the dissemination state to broadcast its ID and its opinion. The dissemination duration is based on its quality of opinion.

The higher the quality of its opinion, the longer it has to broadcast the opinion to its neighbors.

\begin{figure}[tbp]
    \centering
    \includegraphics[width=0.5\linewidth, trim=1.5cm 2cm 1.2cm 1.5cm, clip]{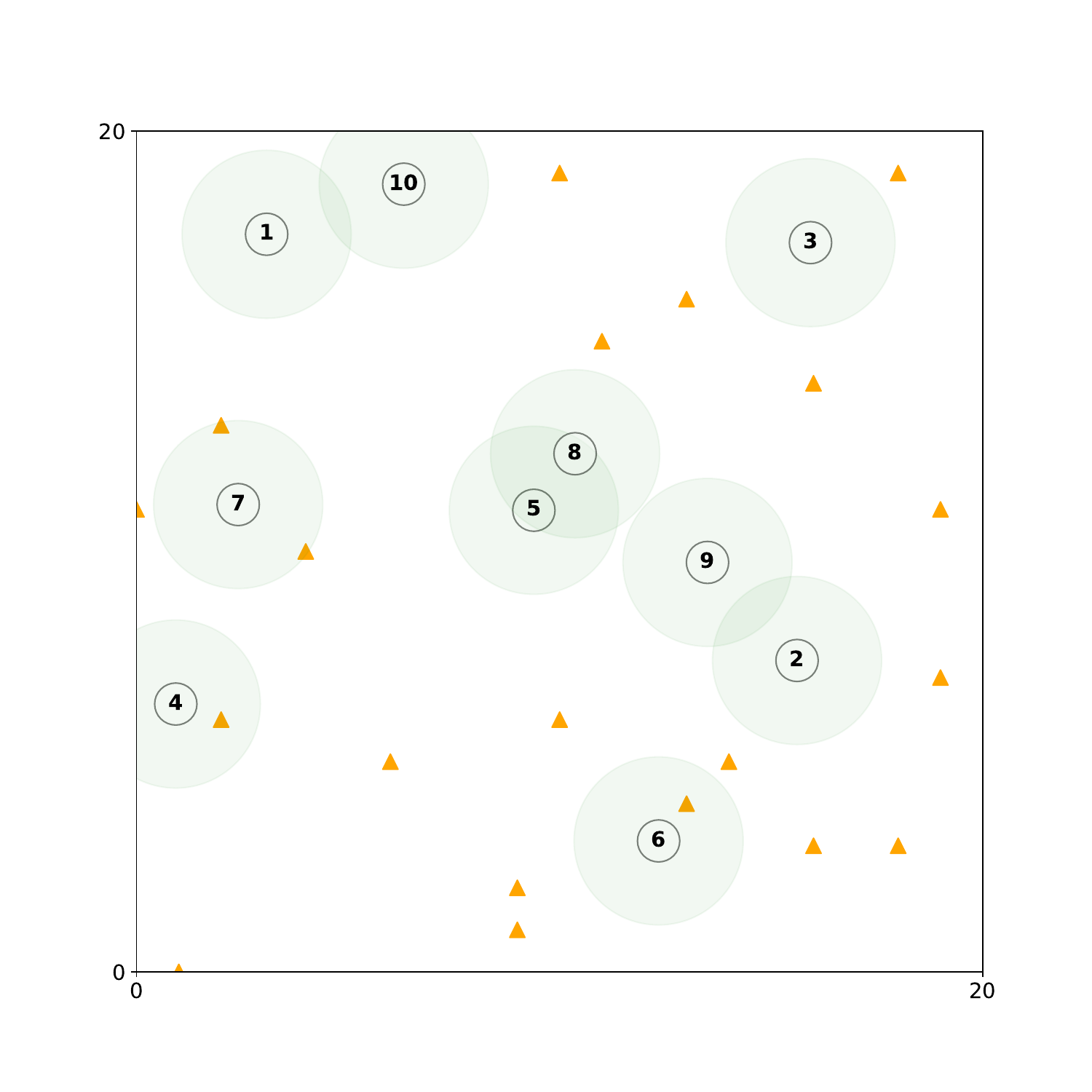}
    \caption{\small Experimental Setup for Distributed Feasibility. Numbers denote robots with the circle shadows indicating the range of observation and orange triangles representing tasks.}
    \label{fig:env_setup}
\end{figure}

During both the exploration and dissemination phases, the robot collects the opinions of its neighbors that broadcast their opinions. At the end of dissemination, the robot counts the collected opinions and chooses the majority opinion. It then goes back to the exploration state with this opinion and begins exploring for new observations.

\begin{figure} [tbp]
\centering
  \begin{subfigure}{0.25\textwidth}
    \includegraphics[width=\linewidth]{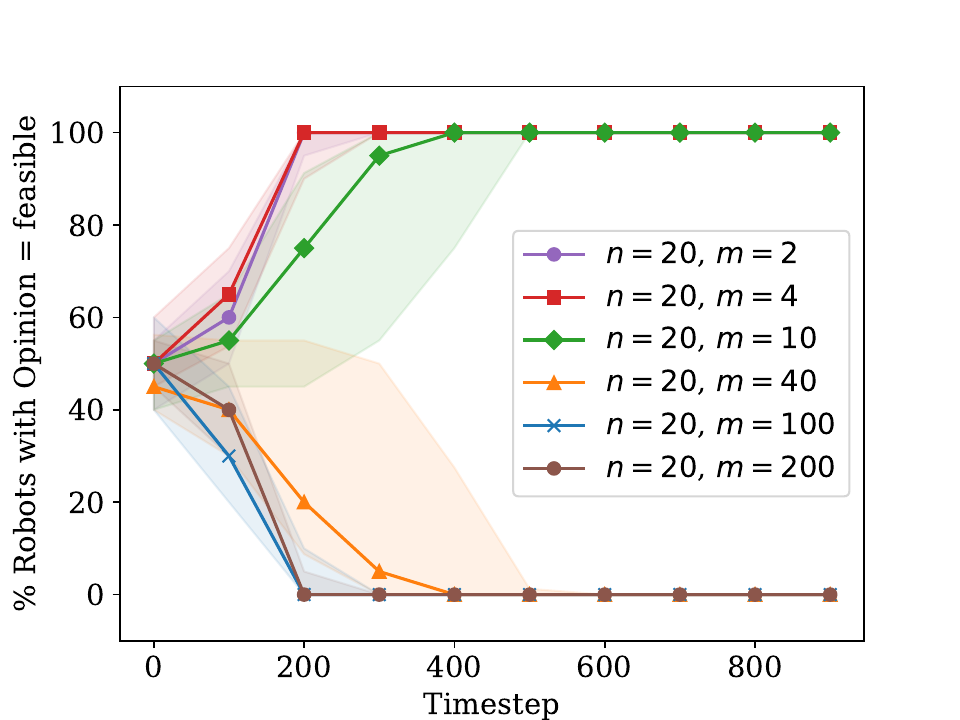}
    \caption{\small Converging Opinions}
    \label{fig:ratio_conv_torus}
  \end{subfigure}%
  \begin{subfigure}{0.25\textwidth}
    \includegraphics[width=\linewidth]{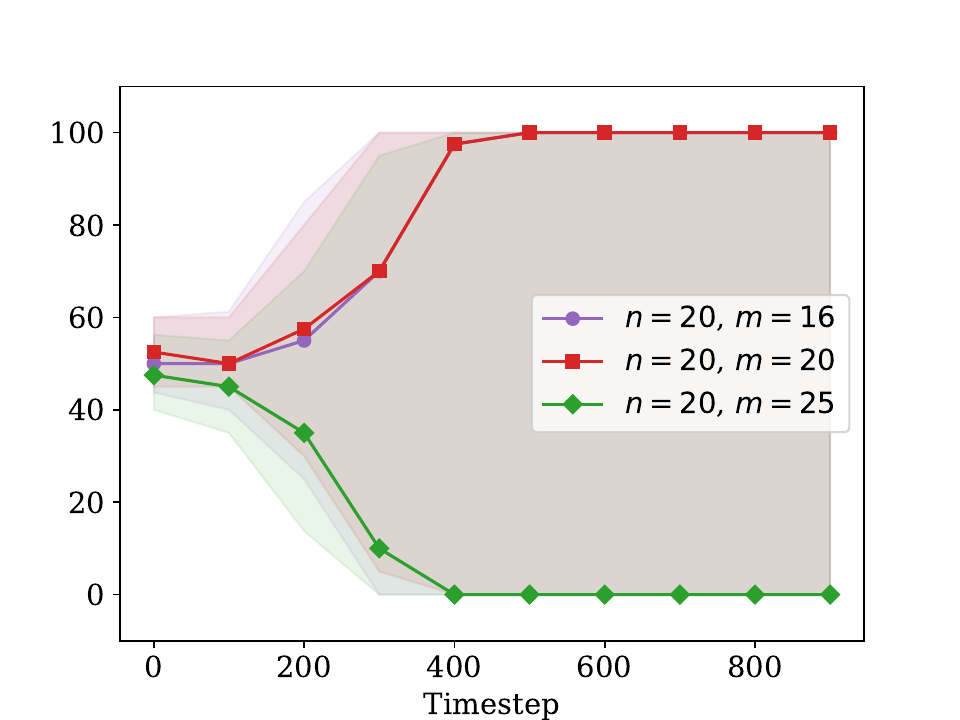}
    \caption{\small Diverging Opinions}
    \label{fig:ratio_div_torus}
  \end{subfigure}
  \caption{\small Percentage of robots holding opinion 1 (feasible) for different numbers of tasks in a collision-free non-unique case. The lines represent the median, and the shadows indicate the upper and lower quartiles for the corresponding colors.}
  \label{fig:ratio_torus}
\end{figure}

\section{RESULTS}

We construct a square environment measuring 20$\times$20 units for our simulation as shown in Fig. \ref{fig:env_setup}. Within this environment, tasks and robots are randomly positioned. Tasks are represented as points and are uniquely identified by their positions. Robots, on the other hand, are represented as circles with a diameter of one unit and possess omnidirectional movement, allowing them to move in any direction at a speed of one unit per time step. Additionally, each robot has an observation range denoted as $d$, within which it can detect tasks and other robots. Each robot is assigned randomly generated ID at the beginning of the simulation. Each robot moves in a straight line for a random duration following an exponential distribution with a specified mean ($\sigma$). It then changes direction randomly and continues moving straight again.  We simulate collective decision-making on feasibility in collision-free movement and non-unique observation. Robots are able to pass each other without impeding their trajectories, and when they reach the boundary of the environment, they continue their movement on the opposite side, resembling a torus configuration space. As for the observation, robots continuously update their observation memory with information about robots and tasks, even if they have previously encountered them.

The number of robots is fixed to $n=20$ robots with an observation range of $d=2$ units, allowing each robot to observe objects and communicate with other robots within this distance. We vary the number of tasks from $m=2$ to $m=200$ resulting in robot-to-task ratios of $\frac{n}{m}=10, ..., 0.1$. Each combination is run for 100 times, and the results are illustrated in  Fig. \ref{fig:ratio_torus}. The y-axis shows the percentage of robots that hold opinion 1 (feasible). 
\begin{figure}[tpb]
    \centering
    \includegraphics[width=0.65\linewidth]{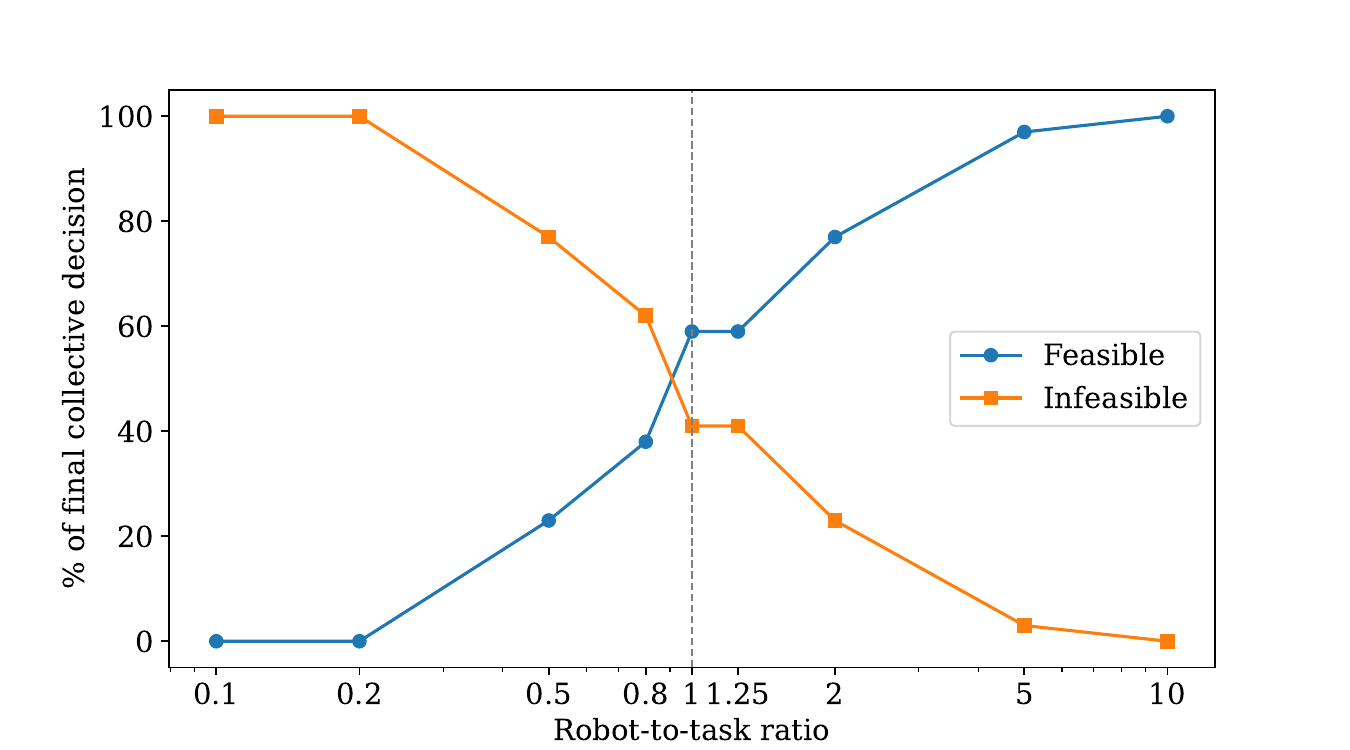}
    \caption{\small Percentage of final collective decision on feasibility with varying robot-to-task ratios, each from 100 runs.}
    \label{fig:dec_torus}
\end{figure}
We observe that when there is a significant difference between the numbers of robots and tasks, the opinions of the robots converge towards the true opinion (see Fig. \ref{fig:ratio_conv_torus}). When the difference between the numbers of robots and tasks is small (robot-to-task ratio close to 1), the opinions of the robots diverge (see Fig. \ref{fig:ratio_div_torus}). The impact of the robots-to-tasks ratio on the final collective decisions on feasibility made by robots can be observed in Fig. \ref{fig:dec_torus}. The final collective decision is made when more than 90\% of the robots hold the same opinion. We observe that the feasible and infeasible decisions are symmetrical, with robot-to-task ratios larger than 2 yielding over 75\% correct feasible decisions (and similarly, ratios smaller than 0.5 yielding correct infeasible decisions). This finding is consistent with the results reported in \cite{Valentini2016,valentini2016b}, suggesting that the DMMD strategy performs effectively when there are substantial differences between the features being considered. Additionally, in Fig. \ref{fig:ratio_conv_torus}, we also see that consensus is reached earlier when the robot-to-task ratio is larger supported by the smaller interquartile ranges (e.g., robot-task-ratio of 10 compared to 2).

\section{CONCLUSION}

The collective decision-making on feasibility in homogeneous robot swarms performs well in a collision-free environment and non-unique observation space, provided there are significant differences between the number of robots and tasks. As expected, faster convergence is observed with a greater difference between the number of robots and tasks, and there is symmetry between feasible and infeasible decisions. In more complex setups, the perception of the ratio may vary in different parts of the arena. For example, a cluster of robots in a corner might have a different consensus on the robot-to-task ratio compared to those that roam freely. Future work will explore collective decision-making on feasibility in more complex scenarios, such as those involving collision avoidance movement and heterogeneity in robots and tasks as well as more complex tasks that require a consensus for assigning robots to tasks.

\end{document}